\documentclass{article}

\usepackage{arxiv}

\usepackage[utf8]{inputenc} 
\usepackage[T1]{fontenc}    
\usepackage{hyperref}       
\usepackage{url}            
\usepackage{booktabs}       
\usepackage{amsfonts,amsmath}       
\usepackage{nicefrac}       
\usepackage{microtype}      
\usepackage{lipsum}
\usepackage{graphicx}
\usepackage{bm}
\usepackage{multirow}
\graphicspath{ {./images/} }

\title{Deep-learning Overloaded Vehicle Identification for Long-span Bridges Based on Structural Health Monitoring Data}

\author{
 Yuqin Li \\
  School of Computer Science and Engineering,\\
  South China University of Technology\\
  GuangZhou, China \\
  \And
  Jun Liu \\
  School of Computer Science and Engineering,\\
  South China University of Technology\\
  GuangZhou, China \\
   \And
 Shengliang Zhong \\
  School of Computer Science and Engineering,\\
  South China University of Technology\\
  GuangZhou, China \\
  \And
  Licheng Zhou \\
  School of Civil Engineering and Transportation,\\
  South China University of Technology\\
  GuangZhou, China \\
  \And
 Shoubin Dong \\
  School of Computer Science and Engineering,\\
  South China University of Technology\\
  GuangZhou, China \\
  \And
  Zejia Liu \\
  School of Civil Engineering and Transportation,\\
  South China University of Technology\\
  GuangZhou, China \\
  \And
  Liqun Tang \\
  School of Civil Engineering and Transportation,\\
  South China University of Technology\\
  GuangZhou, China \\
}

\begin{document}
\maketitle
\begin{abstract}
Overloaded vehicles bring great harm to transportation infrastructures. BWIM (bridge weigh-in-motion) method for overloaded vehicle identification is getting more popular because it can be implemented without interruption to the traffic. However, its application is still limited because its effectiveness largely depends on professional knowledge and extra information, and is susceptible to occurrence of multiple vehicles. In this paper, a deep learning based overloaded vehicle identification approach (DOVI) is proposed, with the purpose of overloaded vehicle identification for long-span bridges by the use of structural health monitoring data. The proposed DOVI model uses temporal convolutional architectures to extract the spatial and temporal features of the input sequence data, thus provides an end-to-end overloaded vehicle identification solution which neither needs the influence line nor needs to obtain velocity and wheelbase information in advance and can be applied under the occurrence of multiple vehicles. Model evaluations are conducted on a simply supported beam and a long-span cable-stayed bridge under random traffic flow. Results demonstrate that the proposed deep-learning overloaded vehicle identification approach has better effectiveness and robustness, compared with other machine learning and deep learning approaches.
\end{abstract}


\section{Introduction}
Nowadays, bridge structural health monitoring (SHM) systems are widely employed to monitor structural responses (strain, displacement, acceleration, etc.) and external excitations (temperature, wind, vehicle loading, etc.) of bridges in order to ensure their safety~\cite{chen2018development}. Among these external excitations, vehicle overloading has become one of the most concerned issues, because it has significant influences on the mechanical performance of bridge structures. A large number of overloaded vehicles in daily transportation may cause fatigue damage, leading to degradation of structures or even failure. Thus, it is of vital significance to develop effective methodologies for overloaded vehicle identification (OVI).

Conventional methods for weighing vehicles include static weighing techniques and weigh-in-motion (WIM) techniques. For static weighing techniques, a static weigh station is utilized to weigh vehicles, which guarantees accurate results. However, these techniques are costly and may cause traffic interruption due to the requirement for vehicles to stop. In contrast, WIM techniques are more widely used in present because they are less time-consuming and less costly~\cite{lu2017bilevel,richardson2014use}. In general, WIM methods can be categorized into pavement-based weigh-in-motion (PWIM) methods and bridge weigh-in-motion (BWIM) methods. For PWIM methods, vehicle weight is measured by devices such as bending plates or load cells, etc., which are embedded into the pavement. Compared to PWIM, BWIM systems have two major advantages~\cite{yu2016state}: (1) better durability of devices is attained, since most sensors are installed under the bridge without direct exposure to the traffic; (2) the installation and maintenance of devices can be implemented without interruption to the traffic.

Nevertheless, BWIM systems are still in face of some drawbacks and challenges, especially for applications for long-span bridges. First, the accuracy of BWIM systems is highly dependent on the accuracy of the predicted influence line which is calculated theoretically~\cite{jian2019traffic}. Nevertheless, it is impossible to predict an influence line that perfectly matches the exact value for a practical simply-supported beam bridge, which brings undesired errors for estimating axle weight. For long-span bridges, the theoretical influence line is much more difficult to simulate accurately in contrast to short-span bridges. Second, for most BWIM systems, the velocity and wheelbase of a vehicle need to be given or determined in advance before the estimation of its weight~\cite{chen2019development}. Usually, the solution is to utilize additional on-the-road axle detectors, such as tape switches and pneumatic tubes, to detect the velocity and wheelbase~\cite{chen2018development}. The use of extra detectors for determining the velocity and wheelbase will undoubtedly leads to higher costs. Third, the accuracy of traditional BWIM algorithms is susceptible to the occurrence of multiple-vehicle presence~\cite{yu2016state}. This is because these algorithms, without the information of real-time positions of every vehicle, have difficulties in separating the contribution of each vehicle from the measured responses~\cite{xia2019infrastructure}. When it comes to long-span bridges in practical engineering, the possibility of the presence of multiple vehicles increases, leading traditional BWIM systems to be more difficult to implement.

To address these challenges, emerging artificial intelligence (AI) technologies show a great potential for providing alternatives to the OVI problem for complex bridge structures. For the OVI task, in fact, the combination of BWIM methods with artificial neural network (ANN), one of the traditional (shallow) machine learning (ML) algorithms, has been investigated~\cite{kim2009vehicle}. ML algorithms have some advantages over traditional methods: (1) domain knowledge, such as the influence line theory that should be utilized in traditional BWIM methods, is not required; (2) dynamic interactions between vehicles and bridges are involved; (3) the effect of multiple-vehicle presence can be considered. However, the performance of ML algorithms strongly depends on the selection of input features or feature engineering. It has been demonstrated that the addition of selected features such as vehicle speed and axle spacing effectively enhances the accuracy of the ANN architecture~\cite{kim2009vehicle}. This indicates that the performance of OVI algorithms depends on not only the establishment of a ML algorithm such as how to select a proper algorithm and train it for a better generalization property, but also feature engineering based on mechanics theories.

As an extension of traditional ML methodologies, deep learning (DL) algorithms use a general-purpose learning procedure to automatically learn good features~\cite{lecun2015deep}, which is one of the key advantages of DL. DL has also been demonstrated to have a prospect in broad applications for civil engineering as well~\cite{rajput2022automatic,ye2019review}. At present, there have been some researches on DL methods for OVI the task. Zhang et al.~\cite{zhang2019methodology} used Faster RCNN network to acquire the vehicle parameters, including the length, number of axles, speed, and the lane that the vehicle is in. Some research work~\cite{ayazi2020data,bosso2020use,lin2020analysis} uses DL methods to explore the factors that lead to vehicle overload, analyze the time and location of frequent occurrence of overloaded vehicles, in order to indirectly detect overloaded vehicles. There are also some studies that directly detect overloaded vehicles with the help of DL. Bernas et al.~\cite{bernas2019vehicle} used linear regression and fully connected neural networks to estimate weight of moving vehicle based on vibration measurements. Zhou et al.~\cite{zhou2021novel} used a deep convolutional neural network to estimate the weight of moving vehicles on the bridge. Wang et al.~\cite{wang2021bridge} improved the ability of the model to predict overloaded vehicles by embedding LSTM and attention mechanism in a Bidirectional RNN network. However, the DL models adopted by the existing research are relatively simple, and the ability to mine information needs to be further improved.

In addition, SHM systems are primarily aimed at damage detection and condition assessment. For example, data from the SHM systems can be analyzed to identify abnormalities in a specific system~\cite{huang2020anomaly}. SHM systems can also be combined with other technologies to build more powerful management systems and evaluate the status of the modules~\cite{valinejadshoubi2019development}. With the rapid progress in data-mining techniques, especially the DL technology, it is possible to estimate vehicle weight with monitoring data from SHM systems~\cite{lin2017structural}. Therefore, based on that SHM systems have been widely applied for bridges nowadays, developing DL methods for the OVI task by using structural response data from SHM systems is an innovative and feasible solution.

Thus, this paper focus on developing a DL algorithm for the OVI task on long-span bridges with SHM data as an alternative solution to the conventional BWIM systems. In this paper, a deep-learning overloaded vehicle identification (DOVI) approach for long-span bridges is proposed by the use of simulated SHM data under stochastic traffic flow. 

The contributions of this paper are as followings:

(1) An end-to-end overloaded vehicle identification framework DOVI is proposed, which require no extra domain knowledge such as the static influence line theory, thereby leading to ease in the application for complex bridge structures and the capability to include dynamic interactions between vehicles and bridges.

(2) Temporal convolutional architecture is employed in DOVI, to learn local time features of input sequence of structural health monitoring data.

(3) Results based on simulation data demonstrate that the proposed approach is effective and applicable to both a simply supported beam and a long-span cable-stayed bridge. 

The remainder of this paper is divided into five primary sections. First, the task definition is explained. Second, the framework of the proposed approach DOVI and the description of the CNN-based DL model are introduced. Third, the bridge models for simulations and dataset establishment are described. Fourth, the results and discussion on the performance of the proposed DOVI method are presented. Finally, detailed summaries and conclusions are provided.

\section{Definition of Overloaded Vehicle Identification Task}

\begin{figure}[h]
  \centering
  \includegraphics[width=0.99\textwidth]{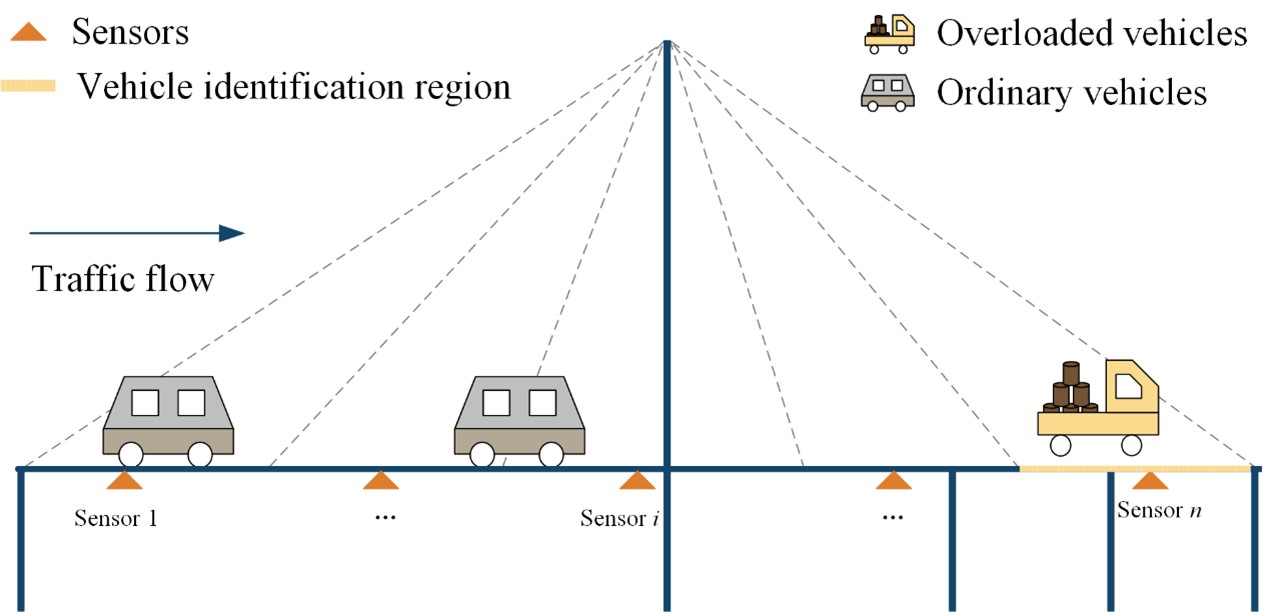}
  \caption{Illustration of a SHM system for bridges under traffic flow.}
  \label{fig:fig1}
\end{figure}

In this section, the definition of the OVI task is introduced. In general, as shown in Figure~\ref{fig:fig1}, SHM systems employ a certain number of sensors to monitor the response of bridges under external excitations such as temperature, traffic, wind, etc. In this paper, the excitations are merely considered as vehicle loads, which are simulated by random traffic flow. The OVI task employs a certain algorithm by use of the measured response of the bridge to identify vehicle weights. Here, several assumptions are adopted in order to simplify the task.

1. Bridge: All bridges in this paper are treated as Euler beams with one-way single-lane roads. In real world, bridges are three-dimensional with two-way multiple-lane roads. However, it is costly and time-consuming to build finite element models for such bridges. Therefore, simplified finite element models are built to simulate structural responses with higher effectiveness. Details of the finite element models of the simply supported beam and the long-span cable-stayed bridge used in this paper will be discussed later.

2. Vehicle: In this paper, the action of one individual vehicle is treated as a concentrated force regardless of the number of axles, considering that vehicles are much smaller in size compared with bridges. Besides, the random traffic flow models with moving forces are applied, which will also be introduced later.

3. Sensor arrangement: Sensors with the number of $n$ are placed evenly on the bridge, as shown in Figure~\ref{fig:fig1}. In this research, bridges are divided into a set of girder sections of the same length along its longitudinal direction. Then, sensors are placed following a simple rule: the number of sensors equals to the number of girder sections and the $i$th sensor is placed in the middle of the $i$th girder section along its longitudinal direction. 

4. Single-label binary classification task: In traditional BWIM systems, vehicle weight is directly estimated. In this paper, the OVI task is treated as a single-label binary classification task. The goal of the task is to identify if an overloaded vehicle exists on the identification region, where the label “true” is assigned if the vehicle weight is larger than a pre-defined weight (which is 30t in this paper). It should be noted that one of the girder sections is selected as the identification region or referred to as the target section. The OVI task is conducted on the target section instead of the whole bridge, because all vehicles will pass through the target section due to the simplification of the one-way single-lane road. Thus, there is no need for multiple identification regions. It should be noted that the length of the girder section has a significant impact on the identification performance, which will be further discussed in this paper.

\section{Framework of DOVI}

\begin{figure}[h]
  \centering
  \includegraphics[width=0.99\textwidth]{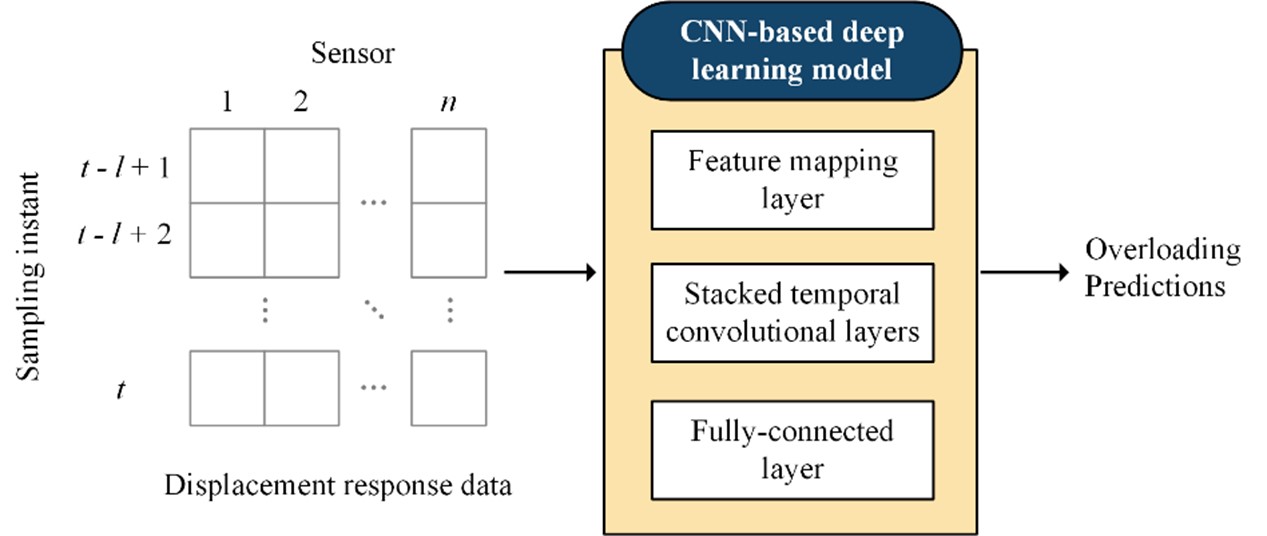}
  \caption{Framework of the proposed approach DOVI.}
  \label{fig:fig2}
\end{figure}

In this section, the detail of the proposed approach DOVI will be presented as below. The framework of the proposed approach DOVI is displayed in Fig~\ref{fig:fig2} Time-series displacement response data from the SHM system are fed into the model as input sequences. Then, the sequences are processed by a CNN-based DL model that consists of a feature mapping layer, stacked temporal convolutional layers and a fully-connected layer. Finally, binary overloaded labels are computed as outputs. In specific, for a given response sequence $X_t$ at time $t$, DOVI classifies whether an overloaded vehicle appears on the target section by the overloaded label $p_t$ at time $t$, which can be presented by

\begin{equation}
    p_t = F(X_t;\omega)
\end{equation}

where $F$ is a nonlinear function established by DL algorithms and $\omega$ is a set of parameters determined during the training phase in a DL model. Note that if a vehicle is identified as an overloaded vehicle, $p_t = 1$, otherwise $p_t = 0$.

The response sequence $X_t$ is described in terms of the response data collected by displacement sensors at sampling instants over a pre-defined past horizon $[t-l+1,t]$ where $l$ is a positive integer. Then, the response sequence is arranged as:

\begin{equation}
X_t = 
\begin{bmatrix}
x_{t-l+1}^1 & \cdots & x_{t-l+1}^i & \cdots & x_{t-l+1}^n \\
\vdots & \ddots & \vdots & \ddots & \vdots \\
x_{t-j}^1 & \cdots & x_{t-j}^i & \cdots & x_{t-j}^n \\
\vdots & \ddots & \vdots & \ddots & \vdots \\
x_{t}^1 & \cdots & x_{t}^i & \cdots & x_{t}^n 
\end{bmatrix}
\end{equation}

where the superscript and subscript denote the sensor number and the sampling instant, respectively, 
$n$ represents the total number of sensors, $x_{t-j}^i,j=0,1,\cdots,l-1$ represents the response measured by the $i$th sensor at sampling instant $t-j$. Besides, the $i$th column of $X_t$, i.e., $x^i = (x_{t-l+1}^i,\cdots,x_{t-j}^i,\cdots,x_t^i)^T$, denote the time-series measurements over $l$ sampling instants for the $i$th sensor. 

The parameters $\omega$ are tuned using training data by the back-propagation algorithm. Consider a training dataset containing $N$ samples, where each sample $v\in N$ is a pairing of a response sequence and a ground-truth overloaded label $p_v^{*}$. The goal of determining the optimal set of parameters $\omega^{*}$ is to minimize the differences between the predicted output $p_v$ and the ground-truth overloaded label $p_v^{*}$ in the training dataset. The optimization procedure of the DL model is to minimize the loss function. In this paper, binary cross entropy is selected as the loss function. Thus, the optimal values of the parameters can be written as

\begin{equation}
    \omega^{*}=argmax_{\omega}\sum_v -[p_v^{*}*log(p_v)+(1-p_v^{*})*log(1-p_v)]
\end{equation}

\begin{figure}[h]
  \centering
  \includegraphics[width=0.99\textwidth]{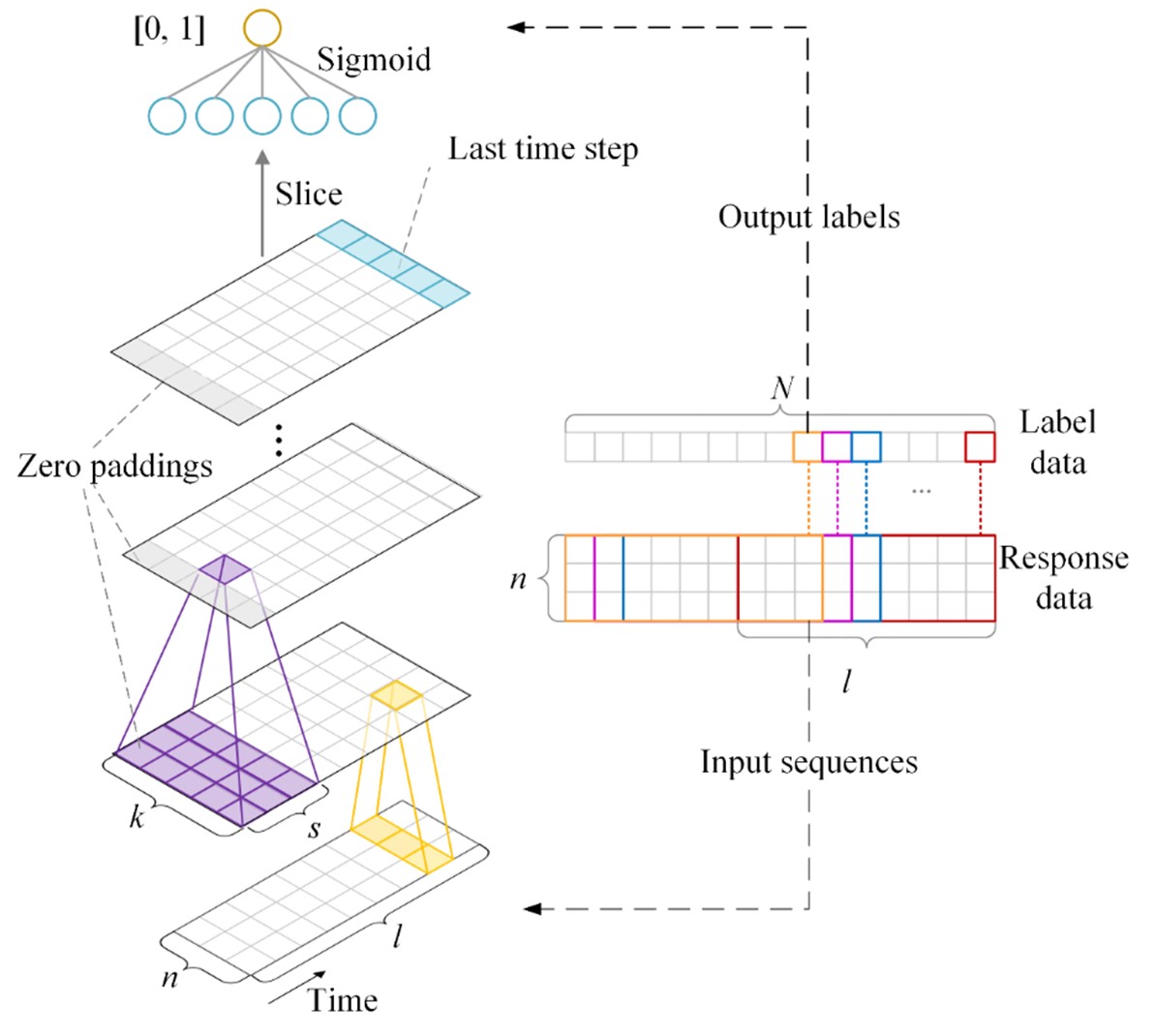}
  \caption{Architecture of DOVI with an illustration of the construction of input sequences.}
  \label{fig:fig3}
\end{figure}

In general, the DOVI consists of a feature mapping layer, one or several stacked temporal convolutional layers and a fully-connected layer, in order to cope with time-series displacement response data. In fact, with regard to process time-series data, RNN and LSTM are common options. However, a recent study conducting a systematic comparison of convolutional and recurrent architectures on sequences modeling tasks indicated that temporal convolutional architectures are more effective than recurrent architectures in various sequence modeling tasks~\cite{bai2018empirical}. Many works have used temporal convolutional architectures to deal with sequence modeling tasks~\cite{torres2021synthesizing}. In this work, stacked temporal convolutional layers are utilized, considering that they are capable of learning local features instead of global features. That is to say, time features of adjacent sampling instant are learned which is intuitively logical because the effect of vehicles on each sensor do not last for a long time. Fig~\ref{fig:fig3} illustrates the model structure along with the construction of input sequences, where 
$s$ is the size of filters, $c$ is the number of stacked temporal convolutional layers, $k$ is the number of filters and $l$ is the length of the sequence (i.e., the length of the pre-defined past horizon. It should be noted that consider input sequences with the length of $l$, $N$ original response data are transformed into $N-l+1$ sample data. In the feature mapping layer, an input sequence is mapped to a high-dimensional feature space by a convolution layer with $s=1$. In stacked temporal convolutional layers, time features are extracted, while both the length of the sequence and the number of features are kept unchanged, as RNN and LSTM do. Finally, the output of the last stacked temporal convolutional layer is then fed into the fully-connected layer, the classifier, to compute a score. If the score is larger than the threshold, which is a hyper-parameter, the classifier assigns a label “true”, i.e., an overloaded vehicle exists on the identification region.

\subsection{Feature Mapping Layer}

In DOVI, an input sequence is first mapped to the high-dimensional feature space. Following the notation above, denote $X_t \in R^{l \times n}$ as the input sequence, $x_t \in R^n$ as a response vector at sampling instant $t$, $f_m \in R^n,m=1,2,\cdots,k$ as a set of filters. Then, the 1-dimensional (1-D) convolution operation computes a feature $\tilde{x}_{t,m}$ at time $t$ by

\begin{equation}
\left\{
\begin{aligned}
    & \tilde{x}_{t,m} = ReLU(f_m * x_t + b_{t,m}) \\
    & ReLU(x) = max(0,x) 
\end{aligned}
\right.
\label{eq:eq4}
\end{equation}

where $b_{t,m}$ is a bias term, the operation $*$ denotes the element-wise product and the rectified linear unit $ReLU$ is an activation function which helps mapping the non-linearity relationship.

The vector $\tilde{x}_t$ after feature mapping at time $t$ and the output $\tilde{X}_t$ of this layer can be written as Eq.~\ref{eq:eq5} and Eq.~\ref{eq:eq6}:

\begin{equation}
    \tilde{x}_t = (\tilde{x}_{t,1},\cdots,\tilde{x}_{t,m},\cdots,\tilde{x}_{t,k}) \in R^k
    \label{eq:eq5}
\end{equation}

\begin{equation}
    \tilde{X}_t = (\tilde{x}_{t-l+1},\cdots,\tilde{x}_{t-j},\cdots,\tilde{x}_{t}) \in R^{l \times k}
    \label{eq:eq6}
\end{equation}

\subsection{Temporal Convolutional Layer}

After the feature mapping layer, the sequence $\tilde{X}_t$ is fed into stacked temporal convolutional layers, where time features are learned. Then $z_{t,s}$, a concatenation of vectors after feature mapping, i.e., $\tilde{x}_{t-s+1},\tilde{x}_{t-s+2},\cdots,\tilde{x}_{t}$, within a window with the length of $s$ can be represented by

\begin{equation}
    z_{t,s} = \tilde{x}_{t-s+1} \oplus \tilde{x}_{t-s+2} \oplus \cdots \oplus \tilde{x}_{t} \in R^{s \ times k}
    \label{eq:eq7}
\end{equation}

It can be seen in Eq.~\ref{eq:eq7} that $z_{t,s}$ only contains information from the past at time $t$ so that no information leaks from future to past. Furthermore, zero padding is added in front of the input sequence to keep the length of the sequence constant as RNN and LSTM do.

Next, let $\tilde{f}_m \in R^{s \times k}, m=1,2,\cdots,k$ refer to a set of filters. Then, a feature $\tilde{t}_{t,m}$ at time $t$ can be computed by the 1-D convolution operation by

\begin{equation}
    \tilde{y}_{t,m} = Activation(\tilde{f}_m * z_{t,s} + \tilde{b}_{t,m})
    \label{eq:eq8}
\end{equation}

The feature vector at time   and the output of this layer can be written as Eq.~\ref{eq:eq9} and Eq.~\ref{eq:eq10}:

\begin{equation}
    \tilde{y}_t = (\tilde{y}_{t, 1},\cdots,\tilde{y}_{t, m},\cdots,\tilde{y}_{t, k}) \in R^k
    \label{eq:eq9}
\end{equation}

\begin{equation}
    \tilde{Y}_{t} = [\tilde{Y}_{t-l+1};\cdots;\tilde{Y}_{t-j};\cdots;\tilde{Y}_{t}] \in R^{l \times k}
    \label{eq:eq10}
\end{equation}

\subsection{Fully-connected Layer}

The feature vector $\tilde{y}_t$ at time $t$ of the output $\tilde{Y}_t$ of the last stacked temporal convolutional layer is fed into the fully-connected layer with sigmoid function as the activation function to compute the score by

\begin{equation}
    \left\{
        \begin{aligned}
            & score = Activation(w^T\tilde{y}_t + b) \\
            & sigmod(x) = \frac{1}{1 + e^{-x}}
        \end{aligned}
    \right.
    \label{eq:eq11}
\end{equation}

where $W^T$ and $b$ are the weight vector and bias term. Finally, a prediction label “true” would be assigned if the score is larger than the threshold, i.e., a vehicle is predicted as an overloaded one.

\section{Model training data Formulation}

\begin{figure}
    \centering
    \includegraphics[width=0.99\textwidth]{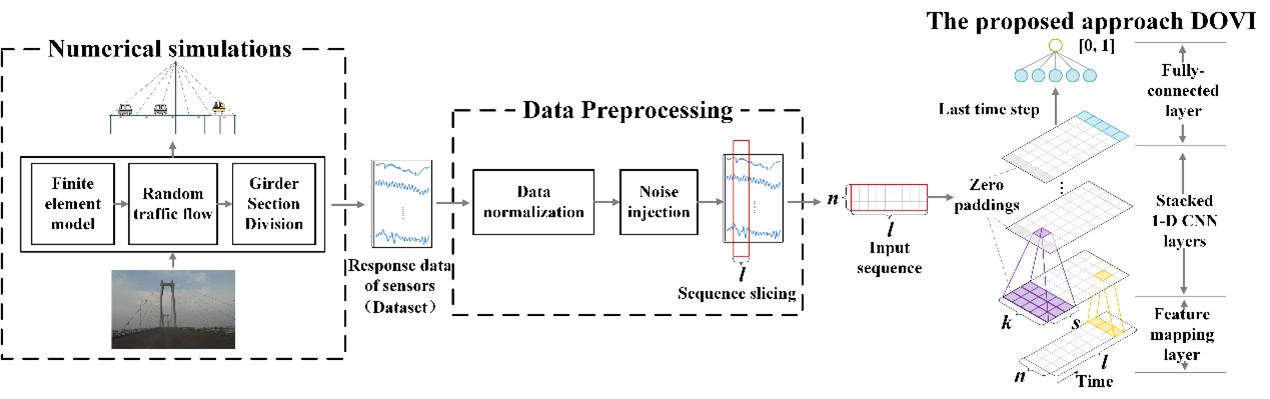}
    \caption{Implementation workflow of the proposed approach DOVI.}
    \label{fig:fig4}
\end{figure}

In this section, numerical simulations and data preprocessing are introduced. As displayed in Figure~\ref{fig:fig4}, the finite element models of the considered bridges and the random traffic flow models are built first, and the bridge simulation model is divided into girder sections. The response data of sensors is collected from the bridge simulation model as the dataset of our experiment. Next, we perform preprocessing such as normalization and noise injection. Finally, the processed data will be used as the input sequence of the DOVI model. 

\subsection{Numerical Simulations of Bridge and Traffic flow}

In order to train and evaluate the proposed approach DOVI, there is a need to establish dataset with simulations. Each dataset consists of two kinds of data: 

1.	Structural response data: Structural response data are generated by the numerical simulations of bridges and traffic flow, where bridges are modeled by the finite element method and excitations are modeled by the random traffic flow. In this paper, the vertical displacement response is utilized.

2.	Binary labeled data: As mentioned above, the OVI task is treated as a binary classification task. Thus, binary labels are decided by the pre-defined weight.

\subsubsection{Finite Element Models of Bridges}

In this research, experiments are conducted on two bridge models: (1) a simply supported beam model (referred to as ‘the simple bridge model’) and (2) a long-span cable-stayed bridge model (referred to as ‘the complex bridge model’), both of which are established by ANSYS.

\begin{figure}
    \centering
    \includegraphics[width=0.45\textwidth]{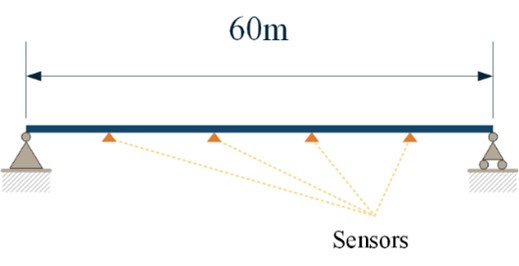}
    \caption{Model of the simply supported beam.}
    \label{fig:fig5}
\end{figure}

In the simple bridge model (SBM), as displayed in Figure~\ref{fig:fig5}, a simply supported T-beam is employed, which is simulated by the BEAM4 girder element in ANSYS. The parameters of the BEAM4 girder element are as follows: the elastic modulus is 33 GPa, the Poisson’s ratio is 0.2, the density is 2600 $kg/m^3$, the section area is 0.94 $m^2$, and the moment of inertia is $I_{yy}=0.23m^4,I_{zz}=0.28m^4$.

\begin{figure}
    \centering
    \includegraphics[width=0.99\textwidth]{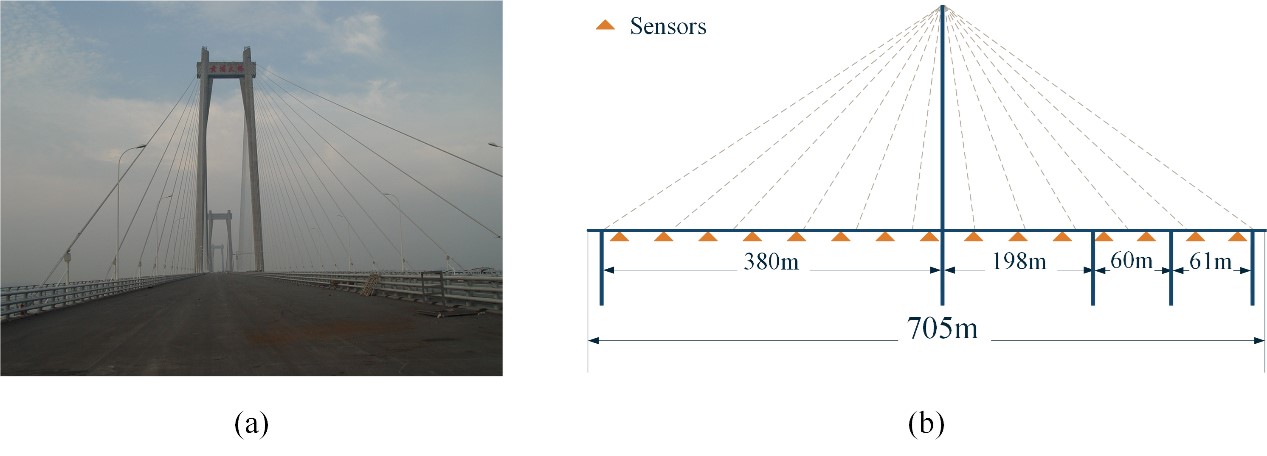}
    \caption{Model of Huangpu Bridge of Pearl River: (a) actual scene; (b) schematic diagram.}
    \label{fig:fig6}
\end{figure}

In this section, details of the complex bridge model (CBM) shown in Figure~\ref{fig:fig6} are presented, which is built based on the north span of Huangpu Bridge of Pearl River located in Guangzhou, China. The bridge is north-south directional with a length of 705m and a deck width of 34.5m. A fish-bone-girder model is built for analysis. Parameters of the model are presented in Table~\ref{tab:tab1}.

\begin{table}
    \centering
    \caption{Parameters of the finite element model of Huangpu Bridge of Pearl River.}
    \resizebox{\linewidth}{!}{
        \begin{tabular}{ccccccc}
        \toprule
        Bridge Component& Element Type& Elastic Modulus (GPa)& Poisson’s Ratio& Density ($kg/m^3$)& Section Area ($m^2$)& Moment of Inertia ($m^4$)\\
        \midrule
        Stay Cables&	LINK10&	200&	0.3&	7850&	various&	/\\
        Main Girders&	BEAM4&	200&	0.3&	7850&	1.72&	/\\
        Fish-bone Girders&	BEAM4&	/&	0&	0&	1& $I_{xx}=1/3,I_{yy}=1/12,I_{zz}=1/12$\\	 
        Tower&	BEAM4&	35&	0.167&	2500&	/&	/\\
        Cross Beams&	BEAM4&	35&	0.167&	2500&	/&	/\\
        \bottomrule
    \end{tabular}
    }
    \label{tab:tab1}
\end{table}

\subsubsection{Random Traffic Flow Models}

The traffic flow follows certain probability distribution in despite of its complexity~\cite{wang2014wind}. In specific, the number of vehicles passing through a certain cross section, the number of vehicles of each type and the corresponding weight of each vehicle type all obey probability distributions~\cite{li1997structural}. In this paper, parameters of these distributions are determined by the statistical data of real traffic conditions on the Huangpu Bridge of the Pearl River, and are employed in the random traffic flow models of two considered bridges. According to reference~\cite{rajput2022automatic}, vehicles are mainly divided into five types. Based on the data recorded at the toll station of the Huangpu Bridge of the Pearl River in July 2016, this paper fits the probability distribution of the vehicle weights of different vehicle types by using the mixed lognormal distribution which can be written as

\begin{equation}
    f(x) = \sum_{i=1}^n w_if_i(x)
\end{equation}

\begin{equation}
    f_i(x) = 
    \left\{
        \begin{aligned}
            & 0,x<0 \\
            & \frac{1}{\sqrt{2\pi}\sigma_ix}exp[\frac{-(\ln{x} - \mu_i)^2}{2\sigma_i^2}],x \geq 0
        \end{aligned}
    \right.
\end{equation}

where $f_i(x)$ represents the probability density function of the $i$th lognormal distribution component with mean $\mu_i$ and standard deviation $\sigma_i$, $w_i$ is the weighting coefficient, satisfies $w_o \geq 0,i=1,2,\cdots,n$ and $\sum_i w_i=1$.

\begin{figure}
    \centering
    \includegraphics[width=0.99\textwidth]{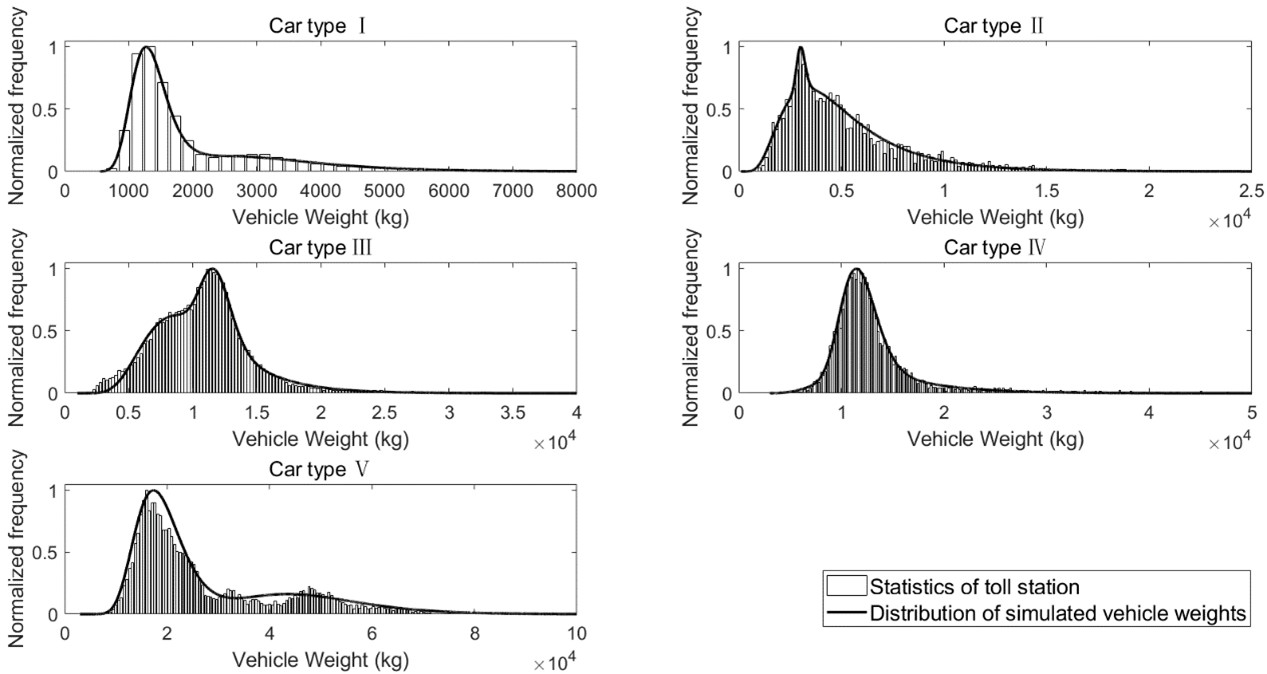}
    \caption{Distribution of vehicle weights of random traffic flow models.}
    \label{fig:fig7}
\end{figure}

The goodness of fit is used to measure the effect of the simulation. Parameters and goodness of fit of the distribution of the vehicle weights are presented in Table~\ref{tab:tab2}. Figure~\ref{fig:fig7} illustrates the distribution of vehicle weights of the random traffic flow models.

\begin{table}[]
    \centering
    \caption{Parameters and goodness of fit of the distribution of the vehicle weights.}
    \resizebox{\linewidth}{!}{
        \begin{tabular}{ccc}
        \toprule
        Car Type&	Parameters&	Goodness of Fit($R^2$)\\
        \midrule
        $\uppercase\expandafter{\romannumeral1}$& $w_1 = 0.337$,$\mu_1=4.970$,$\sigma_1=0.130$;$w_2=0.545$,$\mu_2=7.144$,$\sigma_2=0.211$;$\mu_3=7.883$,$\sigma_3=0.373$& 0.9973\\
        $\uppercase\expandafter{\romannumeral2}$& $w_1 = 0.056$,$\mu_1=8.010$,$\sigma_1=0.670$;$w_2=0.065$,$\mu_2=4.061$,$\sigma_2=0.060$;$\mu_3=8.111$,$\sigma_3=0.543$& 0.9836\\
        $\uppercase\expandafter{\romannumeral3}$& $w_1 = 0.572$,$\mu_1=9.067$,$\sigma_1=0.370$;$w_2=0.293$,$\mu_2=5.815$,$\sigma_2=0.006$;$\mu_3=9.371$,$\sigma_3=0.100$& 0.9932\\
        $\uppercase\expandafter{\romannumeral4}$& $w_1 = 0.134$,$\mu_1=9.390$,$\sigma_1=0.385$;$w_2=0.311$,$\mu_2=9,345$,$\sigma_2=0.149$;$\mu_3=4.936$,$\sigma_3=0.184$& 0.9915\\
        $\uppercase\expandafter{\romannumeral5}$& $w_1 = 0.558$,$\mu_1=9.762$,$\sigma_1=0.256$;$w_2=0.352$,$\mu_2=16.050$,$\sigma_2=0.317$;$\mu_3=10.690$,$\sigma_3=0.258$& 0.98761\\
        \bottomrule
    \end{tabular}
    }
    \label{tab:tab2}
\end{table}

\begin{figure}
    \centering
    \includegraphics[width=0.99\textwidth]{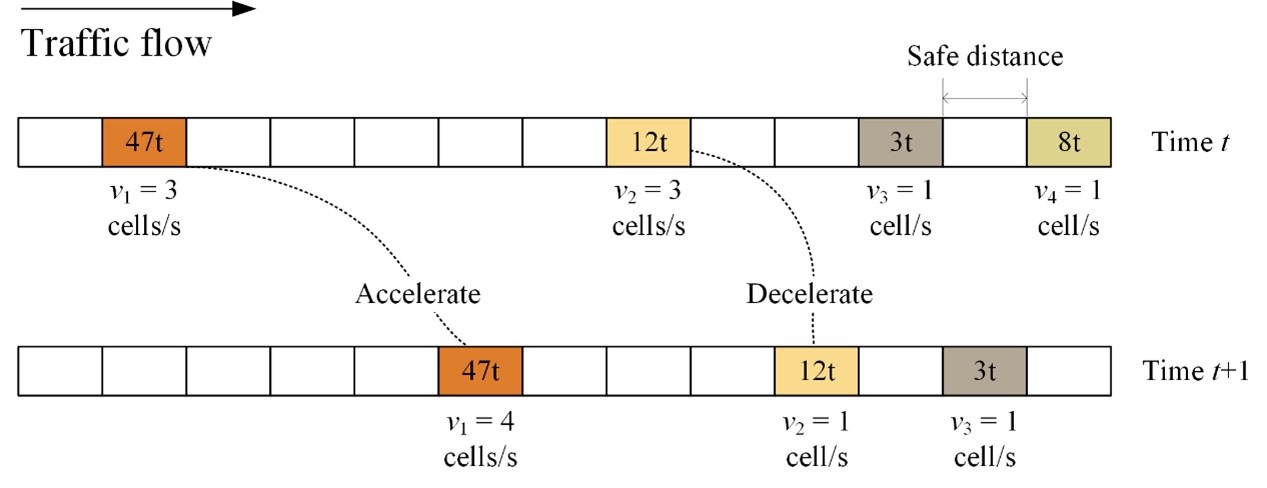}
    \caption{Illustration of the random traffic flow model using cellular automaton.}
    \label{fig:fig8}
\end{figure}

Then, as shown in Figure~\ref{fig:fig8}, the cellular automaton~\cite{nagel1992cellular} is utilized to simulate the simplified one-way single-lane random traffic flow models, where vehicle overtaking and lane-changing are naturally beyond consideration. Vehicles are allowed to accelerate or decelerate following certain rules. From the perspective of the cellular automaton, bridges are considered as a regular grid of cells with the same length. Each cell has its states, including vehicle speed and weight and contains no more than one vehicle regardless of the length of vehicles. 

In this paper, we use Matlab to generate simulation data of the random traffic flow models, which records the load of each cell at each moment. And then, we exert the random traffic flow on the bridge model in ANSYS and extract dynamic vertical displacement response data of the finite element nodes corresponding to the location of sensors. Table~\ref{tab:tab3} exposes various parameters of the random traffic flow models of two experiment bridges.

\begin{table}[]
    \centering
    \caption{Various parameters of random traffic flow models of two bridge models.}
    \resizebox{\linewidth}{!}{
        \begin{tabular}{ccccc}
        \toprule
        Model&	Length of Bridge&	Length of Cells&	Number of Cells&	Maximum Velocity\\
        \midrule
        SBM&	60m&	2m&	30&	4 cells/s\\
        CBM&	705m&	5m&	141&	6 cells/s\\
        \bottomrule
    \end{tabular}
    }
    \label{tab:tab3}
\end{table}

\subsubsection{Girder Section Division}

In this study, bridges are divided into a set of girder sections of the same length along its longitudinal direction. If girder sections are excessively long, there is an increase in the possibility of the simultaneous presence of multiple vehicles on one girder section. This may lead to higher probability of mistakenly recognizing multiple small-weight vehicles as an overloaded one. On the other hand, excessively short sections may result in mistakenly recognizing an overloaded vehicle on an empty girder section due to the influences of vehicles on the neighboring sections. Under ideal condition, when the average number is close to one at each time step, the identification task on a section is similar to the identification task on a single vehicle. 

\begin{figure}
    \centering
    \includegraphics[width=0.99\textwidth]{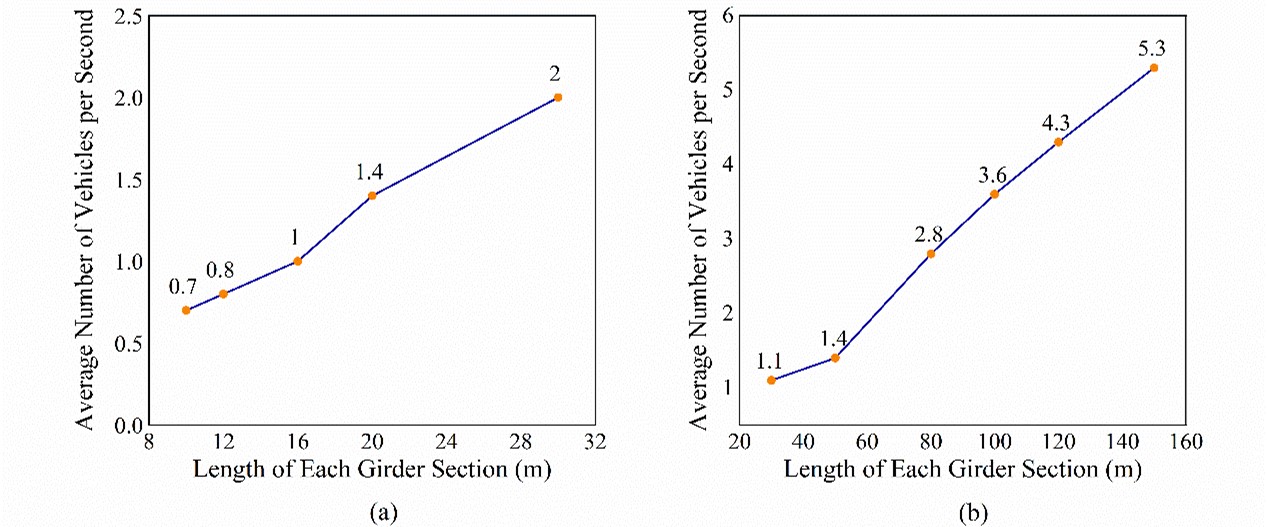}
    \caption{Average number of vehicles on the girder section of different lengths of (a) the simple bridge model and (b) the complex bridge model.}
    \label{fig:fig9}
\end{figure}

Therefore, experiments on the optimal length of the girder section of two bridge models are conducted. Note that the length of each section should be a multiple of the length of cell because girder sections are composed of cells in the random traffic flow model. The experimental results are shown in Figure~\ref{fig:fig9}. Therefore, the simple bridge model finally chooses 16m as the optimal length (the bridge of 60m long is divided into 4 girder sections, where the last girder section is 12m and the others are 16m), while the complex bridge model chooses 50m as the optimal length (the bridge of 705m long is divided into 15 girder sections, where the last section is 5m and the others are 50m).

Finally, use the simulation data of the random traffic flow models to generate binary labeled data: label “1” (i.e., “true”) is assigned if any more-than-30t vehicle (i.e., a cell with the weight of 30t or higher) exists on the target girder section, otherwise label “0” is assigned. 

\begin{figure}
    \centering
    \includegraphics[width=0.99\textwidth]{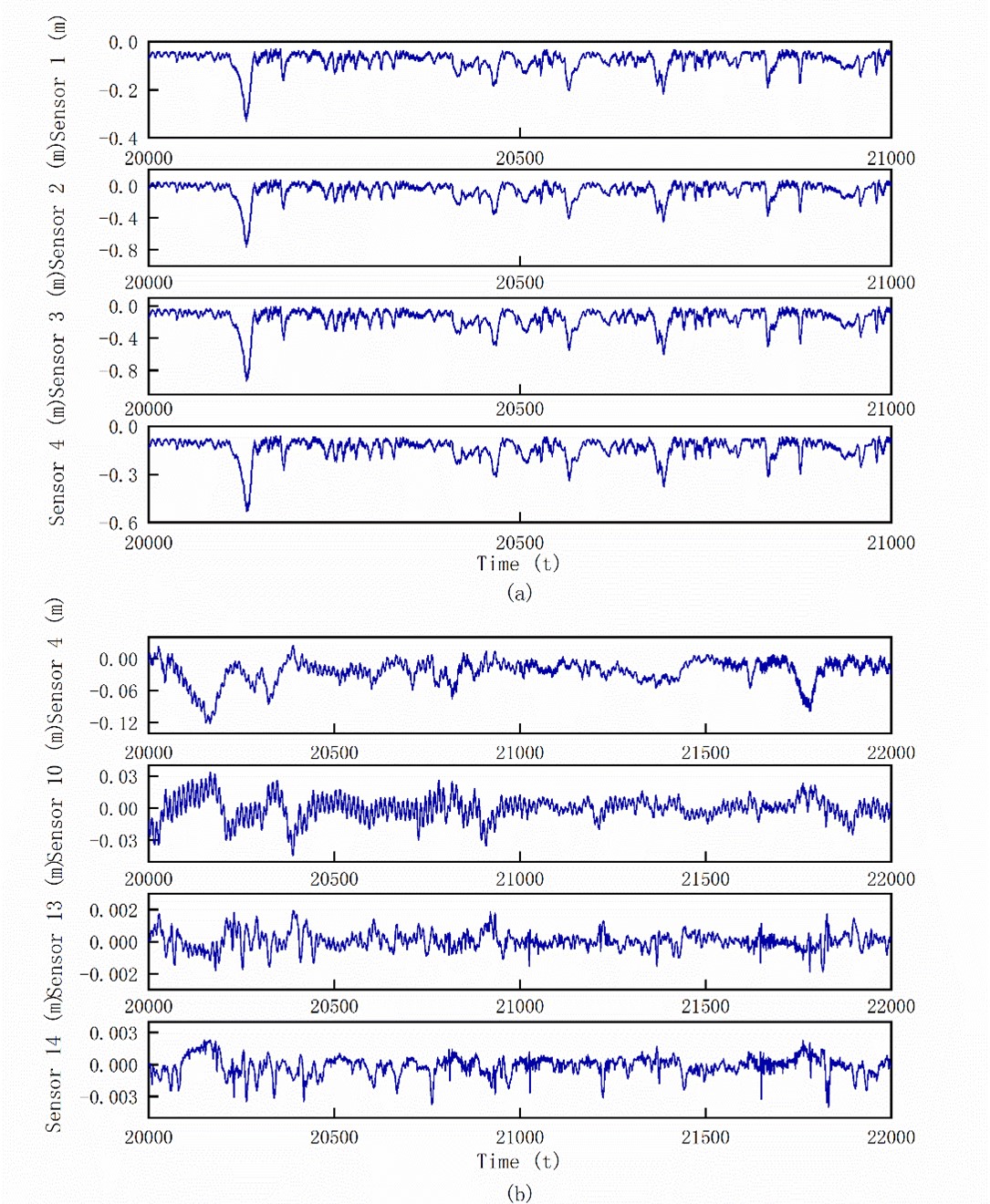}
    \caption{Examples of simulated displacement data of (a) the simple bridge model and (b) the complex bridge model.}
    \label{fig:fig10}
\end{figure}

Table~\ref{tab:tab4} illustrates important parameters of two datasets, where the dataset on the simple bridge model is referred to as “Dataset-SBM” and the dataset on the complex bridge model is referred to as “Dataset-CBM”. Figure~\ref{fig:fig10} shows the displacement response data for several sensors collected over 2000 sampling instants for the two considered bridge models as examples.

\begin{table}[]
    \centering
    \caption{Important parameters of two datasets for training ML models on two bridge models.}
    \resizebox{\linewidth}{!}{
        \begin{tabular}{ccccc}
        \toprule
       Dataset&	Type of Bridge&	Data Number&	Time Step&	Sensor Number\\
        \midrule
        Dataset-SBM&	Simply Supported Beam&	100,000&	0.5s&	4\\
        Dataset-CBM&	Long-Span Cable-Stayed Bridge&	100,000&	1s&	15\\
        \bottomrule
    \end{tabular}
    }
    \label{tab:tab4}
\end{table}

\subsection{Data Preprocessing}

After introducing the numerical simulations (the finite element models, the random traffic flow models and girder section division), the following introduces the data preprocessing. 

\subsubsection{Data Normalization}

Normalization is applied on the sample data, in order to make the data in approximately the same order of magnitude. It should be note that normalization is applied on three dataset individually. The input sample data is normalized to have zero mean and scaled to the range of [-1, 1] as ~\cite{jeong2019sensor}:

\begin{equation}
    x_t^i \leftarrow \frac{x_t^i-\overline{x}^i}{max_t(|x_t^i|)}
\end{equation}

where $\tilde{x}^i$ is the mean of the response data of the $i$th sensor over time, and $max_t(|x_t^i|)$ is the maximum absolute value of the response data for the $i$th sensor.

\subsubsection{Noise Injection}

\begin{figure}
    \centering
    \includegraphics[width=0.99\textwidth]{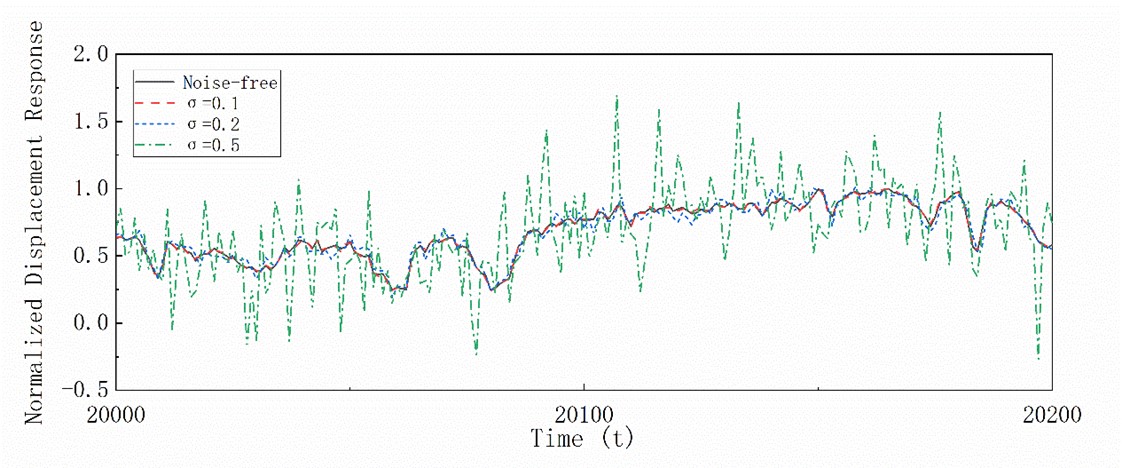}
    \caption{ Example of the response data with noise for Sensor 14 of the complex bridge model.}
    \label{fig:fig11}
\end{figure}

In real-world circumstances, noise is an unavoidable factor. Caused by environmental factors, noise decays the signal by decreasing the signal-to-noise ratio, which may render OVI approaches ineffective in recognizing vital information. Thus, experiments should be implemented under different levels of noise to test the robustness of the model. In this paper, white Gaussian noise is injected into the dataset after normalization by

\begin{equation}
    x_t^i=x_t^i+\xi,\xi \in N(0,\sigma)
\end{equation}

where $\xi$ represents the noise that follows the Gauss distribution with the standard deviation $\sigma \in {0.1,0.2,0.3,0.4,0.5}$, and $\overline{x}_t^i$ represents the normalized response with noise. Figure~\ref{fig:fig11} illustrates an example of response data with noise for Sensor 14 of the complex bridge model.

\subsubsection{Sequence slicing}

Finally, a window of length $l$ is used to intercept the response data to obtain a series of sequence slices. Each sequence slice is an input sample, and the overloaded label at the last time step of each sample will be used as the label of the sample. 

\section{Model Evaluation}

In this section, experiments are conducted on all girder sections of two bridge models for a thorough comparison between DOVI and other popular ML approaches. These approaches include ML algorithms such as logistic regression (LR) and gradient boosting decision tree (GBDT) and DL algorithms such as DNN, RNN and LSTM. 

\subsection{Dataset and Evaluation Metric}

We use simulated dataset to evaluate for experiments. The established dataset is divided into training set, validation set and testing set. The three datasets are divided with the ratio of 6: 2: 2 in dataset size. Specifically, the training set consists of the first 60,000 data, the validation set consists of the data from the 60,001st to the 80,000th data, and the testing set consists of the last 20,000 data.

F1-score (i.e., the harmonic mean of precision and recall) is adopted as the evaluation metric, where precision, recall and F1-score can be presented respectively as 

\begin{equation}
    \left\{
    \begin{aligned}
        P& =& \frac{TP}{TP+FP}\\
        R& =& \frac{TP}{TP+FN}\\
        F& =& \frac{1}{\frac{1}{P} + \frac{1}{F}}
    \end{aligned}
    \right.
\end{equation}

TP is short for “true positive”, i.e., a case where the model makes an accurate identification of an overloaded vehicle. FP is short for “false positive”, which occurs when the model mistakenly recognizes an ordinary vehicle as an overloaded one. FN is short for “false negative”, which may arise when the model fails to identify an overloaded vehicle. 

\subsection{Comparative Investigation of Different Approaches}

Table~\ref{tab:tab5} shows the experimental platform and the software configuration, where the ML package Scikit-learn in Python is employed for the development of LR and GBDT and the DL package Tensorflow and Keras are utilized for the development of DOVI, DNN, RNN and LSTM.

\begin{table}[]
    \centering
    \caption{The experimental platform and the software configuration.}
    \resizebox{\linewidth}{!}{
        \begin{tabular}{ccc}
        \toprule
       Category&	Item&	Parameters\\
        \midrule
        \multirow{2}*{Hardware Configuration} &CPU	&Intel(R) Xeon(R) CPU E5-2603 v4 @ 1.70GHz 6 Cores\\
	   &GPU	&NVIDIA GeForce GTX 1080 Ti\\
        \hline
        \multirow{5}*{System Software and Environment} &Operating System&	Ubuntu 16.04.3 LTS\\
	&Python	&Python 2.7.14\\
	&Scikit-learn &Scikit-learn 0.19.1\\
    &TensorFlow	&TensorFlow-gpu 1.4.0\\
	&Keras	&Keras 2.1.5\\
        \bottomrule
    \end{tabular}
    }
    \label{tab:tab5}
\end{table}

\begin{figure}
    \centering
    \includegraphics[width=0.99\textwidth]{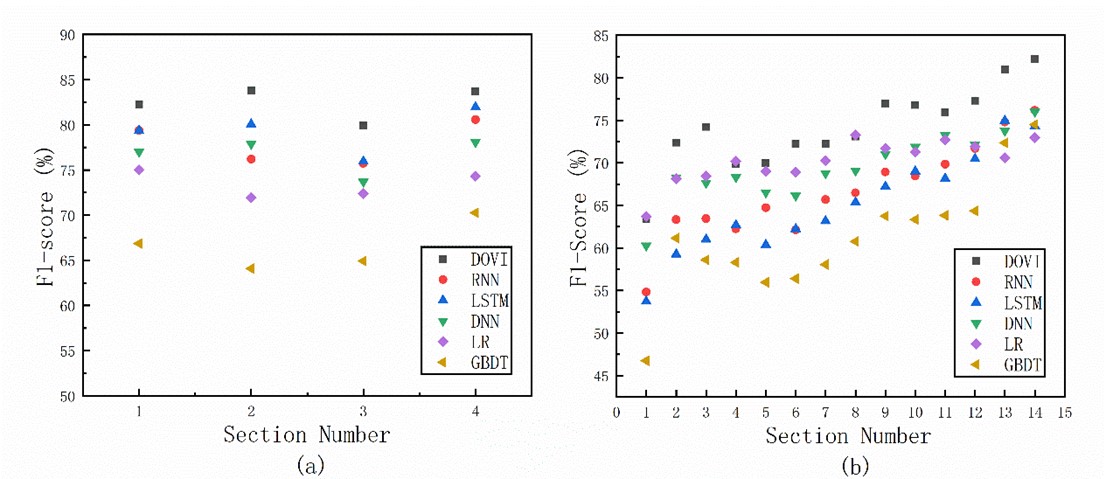}
    \caption{Performance of different approaches on different target sections of (a) the simple bridge model and (b) the complex bridge model.}
    \label{fig:fig12}
\end{figure}

First, the proposed approach DOVI is compared with other approaches. As shown in Fig~\ref{fig:fig12}(a), the proposed approach DOVI outperforms any other approach on all girder sections of the simple bridge model. Table~\ref{tab:tab6} shows the F1-score of each approach on the simple bridge model. DOVI renders the best performance with the highest F1-score of 83.80\% on the second girder section, and presents a significant improvement over other approaches by more than 3.7\% for F1-score. Further observation indicates that DOVI outperforms RNN and LSTM, the common choices for processing time-series data, revealing the DOVI to have a stronger capability of extracting time features.
\begin{table}[h]
    \tiny
    \centering
    \caption{F1-score (\%) of each approach on the simple bridge model.}
    \resizebox{0.6\linewidth}{!}{
        \begin{tabular}{ccccc}
        \toprule
        \multirow{2}*{Approach} & \multicolumn{4}{c}{Target Section}\\
        \cline{2-5}
         &No.1	&No.2	&No.3	&No.4\\
        \midrule
        DOVI	&82.24	&83.80	&79.94	&83.70\\
        RNN	&79.38	&76.20	&75.73	&80.58\\
        LSTM	&79.35	&80.06	&75.95	&81.97\\
        DNN	&77.00	&77.87	&73.70	&78.07\\
        LR	&75.00	&71.94	&72.40	&74.31\\
        GBDT	&66.86	&64.09	&64.92	&70.25\\
        \bottomrule
        \end{tabular}
    }
    \label{tab:tab6}
\end{table}
Next, experiments are conducted on all girder sections on the complex bridge model (except the fifteenth girder section, as mentioned above) to compare the proposed approach DOVI with other approaches in both noise-free and noisy situations. As illustrated in Figure~\ref{fig:fig12}(b), the identification performances of DOVI, common ML algorithms and DL algorithms are compared. Table~\ref{tab:tab7} shows the F1-score of each approach on the complex bridge model. Due to space limitations, only some girder sections with better results are listed. Results indicate that DOVI outperforms other approaches on all girder sections other than the first, fourth, and eighth girder sections. 

\begin{figure}
    \centering
    \includegraphics[width=0.7\textwidth]{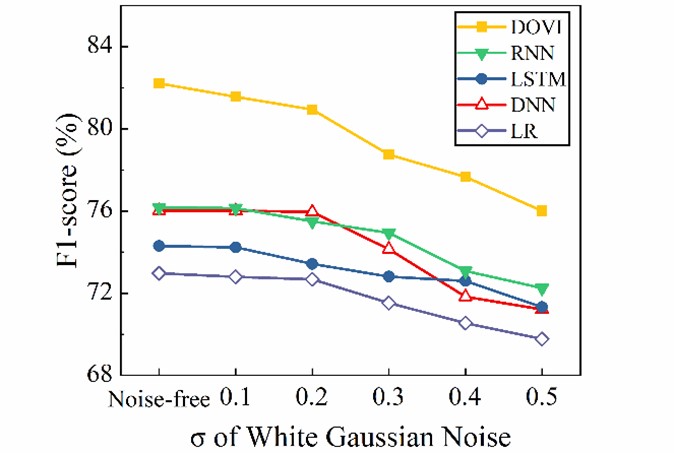}
    \caption{Performance of different approaches on the complex bridge model in noisy situations.}
    \label{fig:fig13}
\end{figure}

Compared with DOVI, DNN is only capable of learning correlations among features at each time step. Thus, from the fact that DOVI outperforms DNN in both bridge models, it can be inferred that time features might contribute to the identification performance. With regard to sequence processing, RNN and its variants such as LSTM are common options. Surprisingly, DOVI outperforms RNN and LSTM in both bridge models, which indicates that recurrent architecture might be less suitable for this task compared to convolutional architecture with the principle of learning local time features. Besides, it can be seen that RNN and LSTM perform well on the simple bridge model but show weak performances on the complex bridge model, while DOVI exhibits stable high performances, which reveals a great robustness of the proposed approach for applications to bridges with different structures.
\begin{table}[]
    \centering
    \caption{F1-score (\%) of each approach on the complex bridge model.}
    \resizebox{0.9\linewidth}{!}{
        \begin{tabular}{cccccccc}
        \toprule
        \multirow{2}*{Approach} & \multicolumn{7}{c}{Target Section}\\
        \cline{2-8}
         &No.8	&No.9	&No.10	&No.11	&No.12	&No.13	&No.14\\
        \midrule
        DOVI	&73.11 	&77.00 	&76.80 	&75.94 	&77.32 	&80.96 	&82.21 \\
        RNN	&66.48 	&68.93 	&68.45 	&69.84 	&71.71 	&74.79 	&76.16 \\
        LSTM	&65.36 	&67.21 	&69.01 	&68.17 	&70.49 	&74.95 	&74.30 \\
        DNN	&69.09 	&71.05 	&71.88 	&73.25 	&72.15 	&73.77 	&76.02 \\
        LR	&73.31 	&71.70 	&71.29 	&72.72 	&71.93 	&70.59 	&72.97\\ 
        GBDT	&60.77 	&63.75 	&63.34 	&63.83 	&64.36 	&72.35 	&74.51 \\
        \bottomrule
    \end{tabular}
    }
    \label{tab:tab7}
\end{table}
In practical engineering, noise is inevitable due to measurement errors of devices and other environmental interferences. Thus, the noise immunity of the proposed approach should be investigated. The 14th girder section of the complex model is selected to be the target section for the reason that all approaches give out their best or second-best performance on it. Table~\ref{tab:tab8} exposes several important hyper-parameters of the optimal DOVI model on the target section. 
\begin{table}[]
    \centering
    \caption{Values of some important hyper-parameters.}
    \resizebox{\linewidth}{!}{
        \begin{tabular}{ccc}
        \toprule
        Name &Value &Description\\
        \midrule
        Sequence Length	&8	&The length of the input sequence of the network.\\
        Stacked Layers	&3	&The number of stacked temporal convolutional layers.\\
        Filter Size	&3	&The size of filters in stacked temporal convolutional layers.\\
        Number of Filters	&64	&The number of filters in convolutional layers.\\
        Learning Rate	&0.002	&The initial learning rate of Adam algorithm.\\
        Batch Size	&128	&The size of data batch used in every training iteration. \\
        \bottomrule
    \end{tabular}
    }
    \label{tab:tab8}
\end{table}
Next, a comparison study between the proposed DOVI method and other approaches is implemented on the selected target section under five levels of the white Gaussian noise, as displayed in Figure~\ref{fig:fig13}. Worse performances can be observed when the signal-to-noise ratio increases for all approaches. As can be seen, the effect of noise on all approaches, except DOVI, is small, when $\sigma$ increases from 0 to 0.1. Then, when $\sigma$ increases from 0.1 to 0.2, only the F1-scores of DOVI, RNN and LSTM decrease obviously. Besides, significant drops in F1-score exist for DOVI, DNN and LR, when $\sigma$ increases from 0.2 to 0.4 while a rapid decrease in F1-score can be seen for RNN when $\sigma$ increases from 0.3 to 0.4. Nevertheless, the highest F1-scores can be obtained by DOVI in contrast to other approaches by a wide margin in all scenarios.

In summary, results demonstrate a better effectiveness and a favorable robustness of the proposed approach DOVI for the OVI task, as compared to other ML and DL algorithms.
\subsection{Analysis on Presence of Multiple Vehicles}

Although the effectiveness of the proposed approach for the complex bridge model has been demonstrated as shown in Figure~\ref{fig:fig12}(b), it is still not clear enough if DOVI performs well when multiple vehicles are present on the long-span bridge. In this section, two questions on this aspect are to answer: (1) When overloaded vehicles are near but not on the target section, the response on the target section is strongly affected by overloaded vehicle on the neighboring section as well as ordinary vehicles on the target section simultaneously. Can DOVI still make a right prediction in this situation? (2) When several vehicles under 30t are on the target section simultaneously and the total weight of these vehicles exceeds 30t, will DOVI gives out a wrong prediction? 

\begin{figure}
    \centering
    \includegraphics[width=0.99\textwidth]{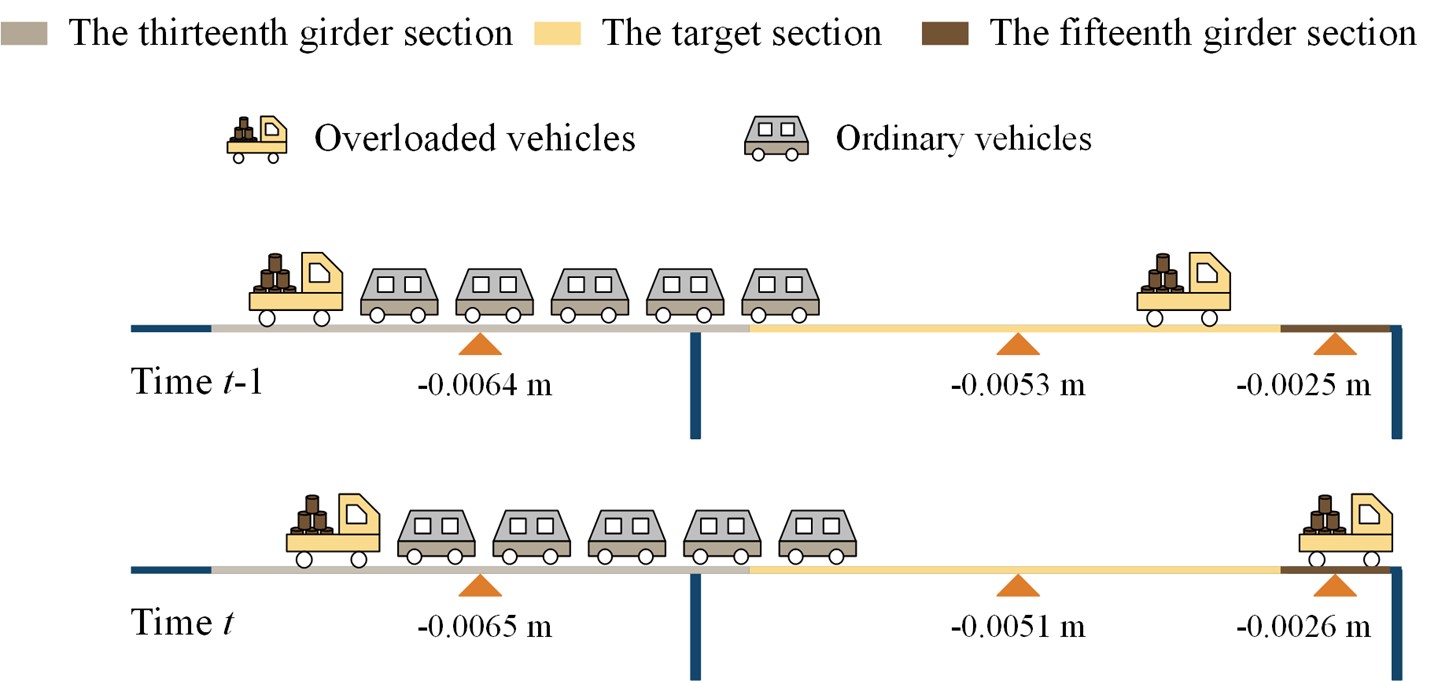}
    \caption{An example of DOVI correct prediction when overloaded vehicles are on the neighboring sections of the target section (the thirteenth and the fifteenth girder section).}
    \label{fig:fig14}
\end{figure}

To answer the first question, we start by counting the number of the cases in the testing set when overloaded vehicles exist on the thirteenth girder section but not exist on the target section. Such cases are 1386 in total. Then, among these cases, 24 false positive predictions are found. That is to say, there are only 1.73\% cases where DOVI mistakenly believes that an overloaded vehicle exists on the target section when the overloaded vehicle is actually on the thirteenth section. This indicates that the proposed approach DOVI renders favorable performance in this kind of situation. Figure~\ref{fig:fig14} demonstrates an example of DOVI making a right prediction when two overloaded vehicles exist on the thirteenth and the fifteenth girder section, respectively. 

\begin{figure}
    \centering
    \includegraphics[width=0.99\textwidth]{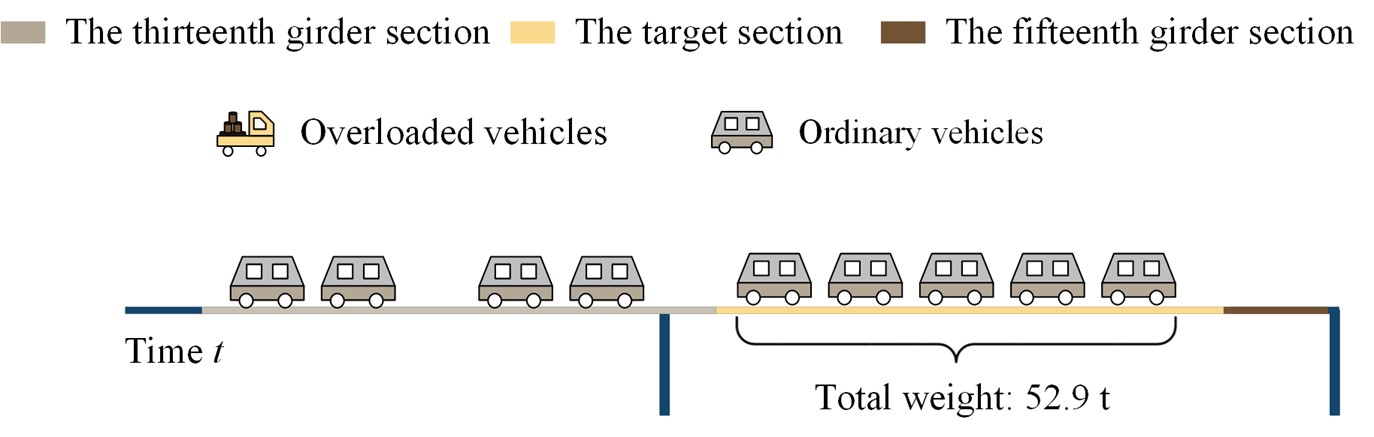}
    \caption{An example of correct prediction when several vehicles (<30t) are on the target section.}
    \label{fig:fig15}
\end{figure}

\begin{figure}
    \centering
    \includegraphics[width=0.6\textwidth]{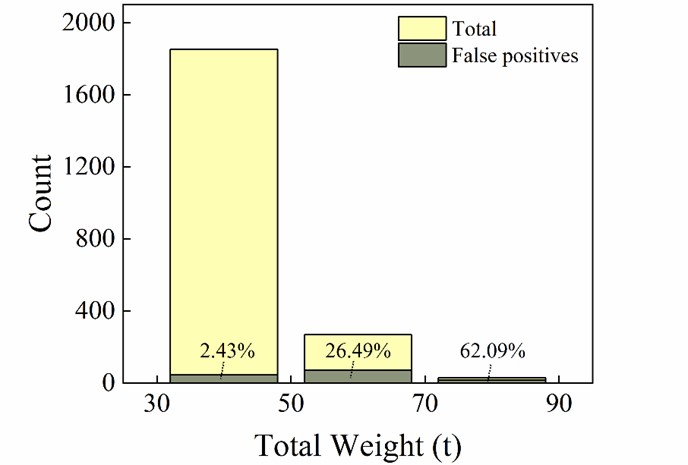}
    \caption{Illustration of distribution of total weights of all vehicles on the target section along with the false positive ratio.}
    \label{fig:fig16}
\end{figure}

The second question is answered by statistical information similarly. There are 2150 cases in the testing set, when no overloaded vehicles exist on the target section but the total weight of ordinary vehicles on this section exceeds 30t. Among these cases, there are only 6.23\% cases where DOVI mistakenly identifies several light vehicles as an overloaded one. An example is presented in Figure~\ref{fig:fig15} that DOVI predicts correctly when multiple vehicles with an individual weight below 30t are on the target section simultaneously. Figure~\ref{fig:fig16} demonstrates the distribution of the total weight of all vehicles on the target section and the corresponding false positive ratio. It can be seen that when the total vehicle weight is lower than 70t, the method seems to exhibit satisfactory performance. 

In conclusion, compared with the traditional BWIM methods, the proposed approach DOVI shows a considerable performance under the presence of multiple vehicles.
\section{Conclusions}

In this paper, a DOVI approach is proposed for long-span bridges by utilizing SHM data to identify overloaded vehicles. In contrast to BWIM methods, DOVI does not require the influence line, vehicle velocity and wheelbase, while the occurrence of multiple vehicles is under consideration. Model evaluations are conducted on the simply supported beam and the long-span cable-stayed bridge under the random traffic flow models. Compared with other approaches based on ML and DL algorithms, results reveal a better effectiveness and a favorable robustness of the proposed approach DOVI approach. Besides, without the need of velocity and wheelbase information as well as domain knowledge such as the influence line theory, DOVI is less costly to be implemented on complex bridge structures. Analysis on the presence of multiple vehicles reveals that DOVI is able to maintain its effectiveness when multiple vehicles presence.

\section{Data Availability Statement}

Some or all data, models, or code that support the findings of this study are available from the corresponding author upon reasonable request.

\section{Acknowledgments}

This work was supported in part by the National Natural Science Foundation of China (Grant Nos. 11972162, 11932007, 12072115, 12072116, 11772131, 11772132, and 11772134) and the Science and Technology Program of Guangzhou, China (Grant Nos. 202102020678 and 201903010046).

\bibliographystyle{unsrt}  
\bibliography{references}

\end{document}